\documentclass{article}
\usepackage{PRIMEarxiv}

\usepackage[utf8]{inputenc} 
\usepackage[T1]{fontenc}    
\usepackage{hyperref}       
\usepackage{url}            
\usepackage{booktabs}       
\usepackage{amsfonts}       
\usepackage{nicefrac}       
\usepackage{microtype}      
\usepackage{lipsum}
\usepackage{fancyhdr}       
\usepackage{graphicx}       
\graphicspath{{media/}}     
\usepackage{float}
\usepackage{amsmath}
\usepackage{makecell}
\usepackage{tabularx}
\title{Marvin: an Innovative Omni-Directional Robotic Assistant for Domestic Environments}

\author{
  Andrea Eirale, Mauro Martini, Dario Gandini, Marcello Chiaberge \\
  Department of Electronics and Telecommunications (DET) \\
  Politecnico di Torino \\
  Torino, Italy \\
  \texttt{\{andrea.eirale, mauro.martini, marcello.chiaberge\}@polito.it} \\
   \And
  Luigi Tagliavini, Giuseppe Quaglia \\
  Department of Mechanical and Aerospace Engineering (DIMEAS) \\
  Politecnico di Torino \\
  Torino, Italy \\
  \texttt{\{luigi.tagliavini, giuseppe.quaglia\}@polito.it} \\
}

\begin{document}
\maketitle

\begin{abstract}
Population ageing and pandemics recently demonstrate to cause isolation of elderly people in their houses, generating the need for a reliable assistive figure. Robotic assistants are the new frontier of innovation for domestic welfare, and elderly monitoring is one of the services a robot can handle for collective well-being. Despite these emerging needs, in the actual landscape of robotic assistants there are no platform which successfully combines a reliable mobility in cluttered domestic spaces, with lightweight and offline Artificial Intelligence (AI) solutions for perception and interaction.
In this work, we present Marvin, a novel assistive robotic platform we developed with a modular layer-based architecture, merging a flexible mechanical design with cutting-edge AI for perception and vocal control. We focus the design of Marvin on three target service functions: monitoring of elderly and reduced-mobility subjects, remote presence and connectivity, and night assistance. Compared to previous works, we propose a tiny omnidirectional platform, which enables agile mobility and effective obstacle avoidance. Moreover, we design a controllable positioning device, which easily allows the user to access the interface for connectivity and extends the visual range of the camera sensor.
Nonetheless, we delicately consider the privacy issues arising from private data collection on cloud services, a critical aspect of commercial AI-based assistants. To this end, we demonstrate how lightweight deep learning solutions for visual perception and vocal command can be adopted, completely running offline on the embedded hardware of the robot.
\end{abstract}


\section{Introduction}
\label{sec:Intro}
In recent years, there has been a significant demographic shift in the global population and, in particular, population aging and its consequences on society need to be seriously taken into account. Indeed, according to the World Population Prospects provided by the United Nations in 2019 \cite{UnitedNations}, life expectancy reached 72.6 years in 2019 and it is forecast to grow to 77.1 years by 2050. In 2018, the amount of persons with an age of 65 or higher reached for the first time the number of children under 5 years. In addition, the United Nations also declared that by 2050 the number of persons aged 65 years or over would overcome the number of youth aged 15 to 24 years \cite{UnitedNations}.
These projections suggest that population aging may drastically affect the entire society, causing social issues in the organization and cost-management of healthcare systems and family units. Moreover, emergency situations such as the COVID-19 pandemic raises critical issues in monitoring isolated people in their houses, which normally need dedicated assistive operators.
Socially assistive robots (SAR) have recently emerged as a possible solution for elderly care and monitoring in the domestic environment \cite{vercelli2018robots}. Although the specific role and objectives of a robotic assistant for elderly care need to be concurrently discussed from an ethical perspective, according to Abdi et al. \cite{abdi2018scoping} diverse robotic platforms for social assistance already exist. These studies often brought researchers to limit their study to the human-machine interaction, realizing companion robots with humanoid \cite{gouaillier2009mechatronic} or pets-like architectures \cite{fujita2001aibo,vsabanovic2013paro}. Such robots have been particularly studied for what concerns dementia, aging, and loneliness \mbox{problems \cite{gongora2019social,gasteiger2021friends}}.
Different studies specifically focus on detailed monitoring tasks, for example, heat strokes \cite{8448739} and fall detection \cite{mundher2014real}.

Besides the healthcare and elderly monitoring purposes, the potential scope of application of an indoor robot assistant is wide, with the enhancement of domestic welfare as a general goal. Indeed, the awareness of air quality risks is rapidly increasing with the spreading of COVID-19 \cite{saini2021sensors}. Moreover, following the Internet of Things (IoT), the paradigm of the house as we know it is changing with the introduction of multiple connected devices \cite{mocrii2018iot}. According to this, recent studies reveal that robots can be identified as complete solutions for future house management \cite{marques2019air}. Robotic assistants represent a promising solution also for monitoring and surveillance in diverse environments such as offices and industrial facilities, with the role of constantly supporting workers while checking potential illness conditions, or accomplishing simple service tasks. 

However, the success of the service assistant robot has not seen a real peak yet: the adoption of these technologies is still limited by the high research focus on technology in the existing prototypes, while a user-centric perspective should guide the design phase. In Section \ref{sec:related}, we discuss the state of the art in assistance robotics and we frame our solution in this context, highlighting its advantages. We believe that the selection of specific target tasks for the robot in a domestic scene is the first necessary step to move toward the spread of robotics assistants as a concrete demonstration of the helpful role of the robot can strongly encourage the user to its adoption. Secondary, we identify a suitable architectural design, flexible and appropriate to the environmental constraints, as the other key factor for a successful final prototype. 

\subsection{Contributions}
\label{subsec:contribut}
In this paper, we propose a novel robotic assistive platform: Marvin. The goal of our mobile robot is to provide basic domestic assistance to the user. More in detail, we identify a set of service functions for the Marvin robot within the overall research scope of socially assistive robots: \textit{user monitoring}, \textit{night assistance}, \textit{remote presence, and connectivity}. In the following sections, we present the layered modular design adopted to conceive Marvin, resulting in a system indifferent to small modifications of the domestic environment and features required by the specific application.
Differently from previously presented robots for home assistance, discussed in Section \ref{sec:related_assistive}, we chose a tiny omnidirectional base platform \cite{doroftei2007omnidirectional}. Indeed, Marvin exploits its restricted footprint and four mecanum wheels to autonomously navigate in a cluttered indoor environment such as the domestic one. Omniwheels and mecanum wheels have already been studied in many \mbox{prototypes \cite{al2018embedded,costa2016localization}} and they are particularly used in industrial robotics applications \cite{qian2017design}, where the flexibility and the optimization of trajectories are a priority. Omnidirectional motion offers a competitive advantage compared to the most commonly employed differential-drive system in unstructured environment navigation. In particular, omnidirectional mobility can be exploited to monitor the user while navigating and avoiding obstacles efficiently. The geometrical asymmetry in the platform's footprint, in conjunction with the omnidirectional capability, can also be exploited to navigate in confined spaces, which are very common in domestic environments. Although the study of human--robot interactions does not fall within the specific scope of this project, we design a telescopic positioning device to adjust the height and tilt of Marvin's camera and its potential user interface. This effectively improves its usability for surveillance purposes offering an extended visual range and facilitating access to its screen for a user-centered visual interaction or telepresence. A custom solution for a positioning system is developed because no commercial device, like a lightweight robotic arm, deals with the strict weight and size requirements dictated by the application. A limitation of this first prototype is that it does not provide manipulation capability, which will be the objective of future studies. 
In addition to an accurate study of the base platform, the Marvin design has been merged with the adoption of state-of-the-art computer vision and AI methods for perception, person tracking, pose classification, and vocal assistance. Deep Learning lightweight models have been selected from recent literature and optimized for real-time inference with the computational embedded hardware mounted on the robot.
Marvin presents an AI-based vocal assistant named PIC4Speech for controlling its actions and selecting the desired task. Differently from commercial solutions such as Google Home or Alexa, the PIC4Speech system described in Section \ref{sec:HRInterface} completely runs offline on the onboard computational device of the robot, avoiding privacy risks and issues of an online cloud-based solution.

Overall, Marvin is a novel robotic solution for domestic and, more generally, indoor user assistance. We distinguish our design choices from existing solutions particularly focusing on service functions in which the mobility constraints dictated by realistic cluttered home environments strongly emerge. To this end, a human-comparable footprint and flexible motion planning, combined with effective visual perception and vocal control systems, can drastically increase the adoption of robotic solutions for home assistance. Therefore, the contributions of this work to robotic assistant research are manifold, and can be summarized as follows:
\begin{itemize}
    \item we conceive a novel modular solution for user monitoring, night assistance, remote presence, and connectivity, prioritizing the agility and flexibility of the platform in complex domestic environments, by adopting a tiny human-comparable footprint and an omnidirectional base platform for the robot (Sections \ref{sec:marvin_concept} and \ref{sec:low_layer});
    \item we design a controllable telescopic positioning device (Section \ref{sec:low_layer}) for easy access to the visual interface and to extend the visual range of the robot;
    \item we develop a real-time AI-based vision system (Section \ref{subsec:CV}) to constantly check potential critical conditions of the user based on their pose, and automatically set up an emergency call;
    \item we propose the PIC4Speech vocal control system (Section \ref{sec:HRInterface}) to provide a reliable, offline vocal interface for the user to express commands to the robot easily.
\end{itemize}

\section{Related Works}
\label{sec:related}
In this Section, we present an overview of the state of the art in assistive service robotics, comparing the most popular architectural solutions proposed in the literature so far, and discussing their points of strength and weakness that led us to design our platform and its sub-components. In the Section \ref{sec:related_assistive}, we firstly present the main assistive robotics platforms presented in the literature, highlighting their peculiar design characteristics. Then, in Sections \ref{sec:related_visual} and \ref{sec:related_vocal}, we briefly introduce the principal technologies that have been used to develop functionalities of Marvin.

\subsection{Assistive Service Robots}
\label{sec:related_assistive}
\begin{figure}[H]
\centering 
\includegraphics[width=15 cm]{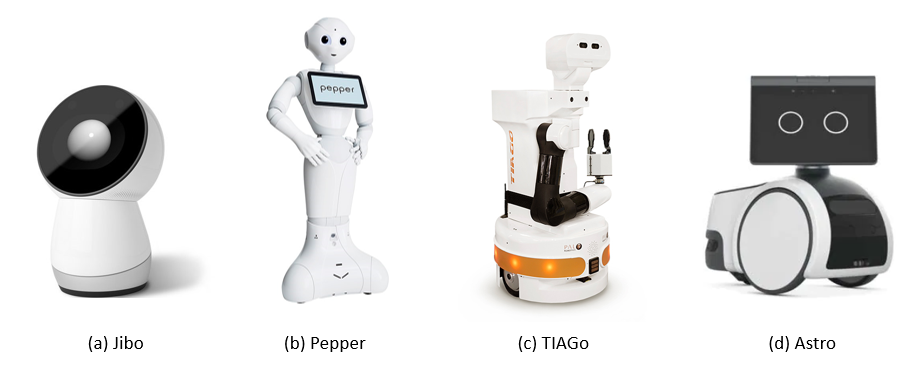}
\caption{Commercial robots developed or suitable for service home-care applications. 
\label{fig:assistive_platforms}}
\end{figure}

In the last years, the robotics research community is focusing its effort on the study of an effective design for an indoor assistant, and different proposals have recently emerged.
Diverse researchers based their study on the human-machine interaction, realizing humanoid companion robots such as NAO \cite{gouaillier2009mechatronic} or pets-like architectures such as Aibo \cite{fujita2001aibo} and PARO \cite{vsabanovic2013paro}. These kinds of robots have been particularly studied for research on dementia, aging, and loneliness problems \cite{gongora2019social,gasteiger2021friends}, although their usage cannot be extended to home assistance without a mobile platform.
Different studies specifically focus on detailed monitoring tasks, for example, heat strokes \cite{8448739} and fall detection \cite{mundher2014real}. However, although their usual expensive cost, they often result to be unused for a long time horizon due to the complex healthcare task they try to accomplish. Indeed, for the pure purpose of a companion robot, marginal differences exist with the more competitive commercial vocal assistants like Alexa, with a much lower cost. Jibo \cite{jibo} (Figure \ref{fig:assistive_platforms}a) is another example of a social robot for the home which falls in this category. Hence, an assistive domestic robot should go beyond the conversational skills of common vocal assistants and we decided to choose a mobile platform, trying to identify a clear, helpful role for the robot in a domestic scenario.

There are already a good amount of research prototypes and few commercial mobile platforms for home-care robotics today. Among them, the HOBBIT robot \cite{fischinger2016hobbit} and the Toyota Home Service Robot (HSR) \cite{hashimoto2013field} are the results of different research projects and they present a similar architecture composed of a wheeled main body equipped with manipulators for grasping objects. The Pepper robot (Figure \ref{fig:assistive_platforms}b) is one of the most popular humanoid robots and it has been also used for nursing and rehabilitative care of the elderly \cite{tanioka2019nursing}. TIAGo \cite{tiagoPAL} (Figure \ref{fig:assistive_platforms}c)is another comparable platform developed for robotics research groups and in general for indoor applications. Even though they are standard differential drive wheeled platforms, all these robots aim at reproducing a human-like overall shape and presence. However, a large footprint and a standard steering system represent strong disadvantages to navigating in a realistic cluttered household environment. The same limitations hold for the SMOOTH robot \cite{juel2020smooth}, the resulting prototype of a research study that aims at developing a modular assistant robot for healthcare with a participatory design process. Three use cases for the SMOOTH robot have been identified: laundry and garbage handling, water delivery, and guidance. In agreement with the authors of the SMOOTH robot, we decided to avoid a robotic arm on the robot, due to the higher control complexity it requires and stability issues it causes when mounted on a tiny lightweight mobile platform. Instead, we designed an innovative controllable positioning device to lift the camera point of view to a reasonable adjustable height, and to allow the user to access the robot visual interface easily. The abilities to carry objects without a manipulator and offer physical support to elderly people are the advantages of the SMOOTH robot design. However, we consider the tiny size and the omnidirectional motion of the platform the crucial design choices to enable the introduction of robot assistants in real-world domestic environments on a large scale.
We discussed that the current state of the art in mobile assistive robots suggests a wide variety of potential configurations for the platform. Compared to previous works, we take into account the following key considerations to enhance the success of robots for domestic applications: 
\begin{itemize}
    \item a \textbf{{tiny} size} 
 fits better in a cluttered domestic environment and improves the acceptance of the presence of the robot in the house;
    \item an \textbf{agile and flexible mobility} guarantees better performance in navigation tasks with complex obstacles and narrow passages;
    \item a \textbf{positioning device} \textls[-25]{for camera and tablet is preferred to a robotic arm for this prototype}.
\end{itemize}

Recently, our design considerations have been strictly confirmed by the lastly emerging commercial proposals. Indeed, in addition to research projects, Amazon has recently launched its commercial home assistant Astro \cite{astroAMAZON} (Figure \ref{fig:assistive_platforms}d). Even though it is still in an experimental stage, Astro can surely be considered an enhanced design thought for end-users, which can visually recognize people and interact with them through a visual interface that aims at conveying expressive reactions and thanks to the Alexa vocal assistant. Robotic platforms such as Astro aim at totally managing the house, also providing surveillance and telepresence services. We can notice that Astro presents a reduced size compared to the typical humanoid platforms to guarantee agile movements in the house and does not represent an oppressive figure for the users, at the cost of not being able to carry items. Moreover, it does not present robotic arms for manipulating objects. With our omnidirectional platform, we aim to improve the mobility and obstacle avoidance of a platform like Astro, together with offering an extended visual range and an easy access to the visual interface thanks to our positioning device. Moreover, Amazon designed the robot to integrate it with the home automation system, exploiting Alexa as a vocal interface. However, this choice exposes Astro to high privacy risks and issues, handling both vocal and visual data of domestic private environments. To prevent such risks, we consider the idea of developing a basic \textbf{offline vocal assistant}, as discussed more in detail in \mbox{Sections \ref{sec:related_vocal} and \ref{sec:HRInterface}.}

\subsection{Visual Perception}
\label{sec:related_visual}
Visual perception in robotics plays a major role, enabling a thorough and detailed scene understanding with a wide variety of visual tasks. Pose estimation \cite{toshev2014deeppose,cao2019openpose}, object detection \cite{zhao2019object}, and semantic segmentation are the main visual processes that allow the robot to perceive and interpret what surrounds it.
The drastic increase of robotic devices both in manufacturing and management facilities and, more recently, in populated environments such as offices, hospitals, and houses, is strictly tied to the breakthrough of Artificial Intelligence (AI) and Deep Neural Networks (DNN) for computer vision applications. From the publication of the ImageNet dataset \cite{deng2009imagenet}, the escalation of Deep Learning from AlexNet to the most recent DNN architectures \cite{alom2018history} is still in progress worldwide.
In particular, the real-time detection of humans is a growing computer vision process that provides support in the application field of surveillance and monitoring, and most critically, allows robots to be placed in populated environments. In particular, person detection is the pillar of every visual-based human--robot interaction. 
An effective perception system of humans is, therefore, a necessary condition to let robots safely plan their activity. According to this, the robotics research community is focusing its effort on person-aware autonomous navigation algorithms \cite{mateus2019efficient}, and the detection of human figures in the visual stream is the first step towards this goal.
Classic deep learning-based one-stage object detectors such as YOLO \cite{redmon2016you} and SSD \cite{liu2016ssd} networks provide the robot with information about the presence of an object in its field of view. However, these approaches only give such information in the form of a bounding box, namely a region of interest of the image where the object is detected. For robotic tasks, especially for human-aware applications, it is crucial to have some additional knowledge about the person. For this reason, pose estimation models represent a more suitable and powerful choice, and their usage is twofold: to detect the presence of humans and provide information about their pose status. State-of-the-art models for human pose estimation \cite{cao2019openpose,PoseNet} provide a skeleton schematic graph of the person. We use PoseNet to estimate human poses as we need a fast inference system and fine-grained information about the human pose. We build our complete pose classification pipeline training a simple DNN, which receives as input key points estimated by PoseNet. Our visual perception pipeline combines both RGB and depth images to autonomously detect emergency conditions of the user, constantly checking if they are standing, sitting, or laying, and to track the person's relative position to enable the human-centered navigation of the robot.

\subsection{Vocal Interface}
\label{sec:related_vocal}
The study of Human--Robot Interface (HRI) has become today a fundamental component for the spreading success of robots in society, allowing the end-users to interact with the robot in different ways. Visual and vocal interfaces are the most common choices to let humans easily interact with a robotic machine. Joysticks and touch screens are other solutions more diffused in the research world during the development phase. In this work, we focused on a vocal interface, to allow the user to call and interact with the robot also from a suitable distance, without the need to access its screen, and to facilitate access to a high-tech device for the elderly. In the last decade, speech processing has seen huge steps forward \cite{moore2015talking} according to the progress of robots and AI. However, training Deep Learning models for vocal assistants requires an extremely high amount of data \cite{skantze2021turn}, that only giants of the market such as Google and Amazon can collect and exploit easily. Moreover, state-of-the-art models in Natural Language Processing (NLP) \cite{tenney2019bert} provide great performances at the cost of a much higher computational cost, which forbids their usage on embedded devices with constrained hardware resources. Commercial solutions such as Siri, Alexa, or Google Home exploit a cascade activation pipeline of multiple models that transfer the computation from the physical device, when triggered, to the cloud servers to run their NLP algorithms. Indeed, the development of a full pipeline of algorithms for fast-interference low-cost vocal assistance in robots is rare to be found in the research literature, although it is a fundamental aspect of human--robot interaction. 
According to this, we decided to kick off a research project, the PIC4Speech vocal assistant, with the aim of providing a low-cost, efficient solution to be executed on board the robotic platforms without the need for expensive hardware and, above all, without relying on a stable internet connection. This last consideration is born from the objective to avoid the exposure of private data to the internet, protecting the development of the robot assistant Marvin from privacy issues, which have been faced by other previous prototypes such as the Astro robot. The complete description of PIC4Speech architecture is reported in Section \ref{sec:HRInterface}. Besides the commercial online assistants, no other similar systems have been identified for a direct, thorough comparison.
PIC4Speech is intended to be an offline vocal assistant for human--robot interaction, even though in this primitive shape of development, its goal is principally to allow the user to give commands to the robot vocally and not to carry out a meaningful social conversation. Indeed, Marvin is considered a robot assistant more than a companion robot and social aspects of communication are not treated in this study.

\section{Requirements}
Against the described state-of-the-art scenario, the researchers at Pic4SeR Center (Interdepartmental Centre for Service Robotics) of the Politecnico di Torino, in association with the researchers at Officine Edison, developed a personal assistant mobile robot called Marvin. The robot has been conceived as a proof of concept to explore the possibilities of autonomous assistive robots in domestic environments, designed for people owning reduced motility, like elderly or people with disabilities. To such an aim, the robot must be able to perform the following service functions:

\begin{enumerate}
    \item \textit{User Monitoring}: the robot should be able to detect a potentially dangerous situation for the user and call for help. To detect a sudden illness or a drift in the person's behavior, many approaches are possible. For example, it is possible to structure the environment with proper sensors, provide the user himself with wearable devices, or exploit visual and perception sensors directly mounted upon the robotic assistant. The first two solutions require intervention in the users’ houses, or the users themselves, which can be perceived as invasive measures in the living environment. Therefore, the monitoring of the person by the robot turns out to be more feasible and immediate. This task requires the robot to track the user as they move within the environment continuously. From a mobility point of view, this also implies the ability to move and reorient the robot sensors at any instant
    \item \textit{Night Assistant}: one of the most critical moments in the daily life of elders is the night-time bedroom-to-toilet journey. The robot should cater to assist in all those situations in which, for whatever reason, the user was unable to reach the electric lighting. The Night Assistant service proposes to accompany the user in any desired location of the domestic environment, enlightening the path and monitoring their movements, giving alarms in the case of need. Again, such a feature requires as-high-as-possible maneuverability to contemporaries providing light and not hampering the user path
    \item \textit{Remote presence and connectivity}: the robot must be provided with the ability to access commonly used communication platforms (e.g., Skype, Whatsapp). This task implies the robot should be able to approach the communication interface for the user to answer or perform calls/video calls. Such ability infers with usability requirements: the human--robot interface needs to be positioned and oriented towards the user
\end{enumerate}

\begin{figure}[H]
\centering 
\includegraphics[]{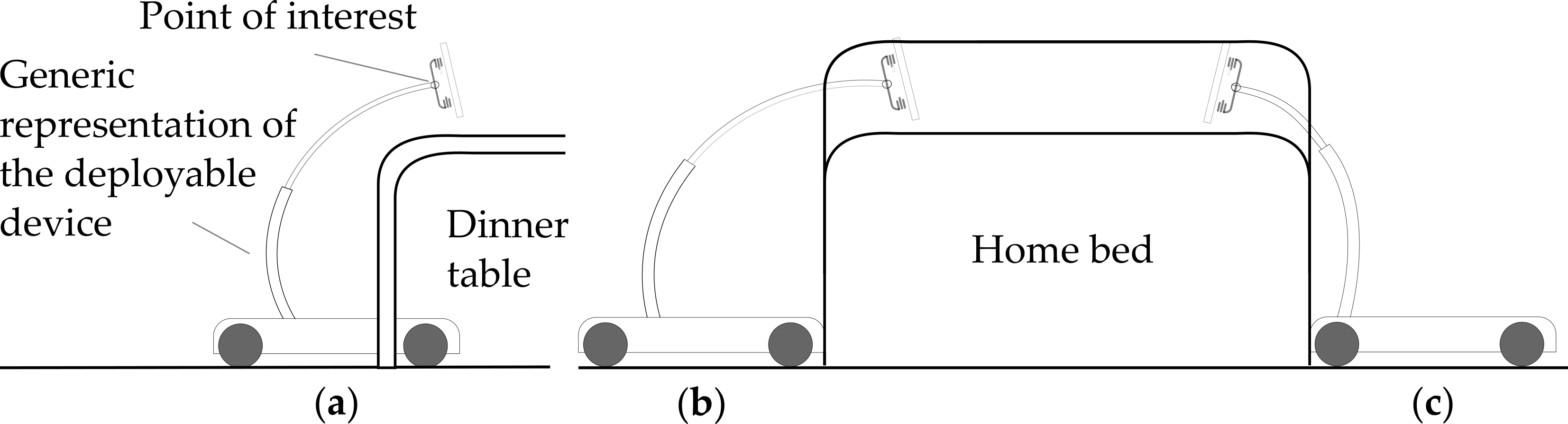}
\caption{Schematic representation of the system in three situations: (\textbf{a}) table approach with under-motion allowed, (\textbf{b}) bed approach with the positioning device mounted on the opposite side relative to the deployment direction, (\textbf{c}) bed approach with the positioning device mounted on the approach side.
\label{fig:PosDev_MountingStr}}
\end{figure}

These tasks, addressed to provide a service to the user, in turn require a series of robotic capabilities. First of all, to properly monitor the user, the system should be able to perceive them, recognize a laying posture and ask for help if necessary. This comports the implementation of pose detection and classification neural networks and external communication via a mobile connection. The monitoring service would become pretty limited without a continuous track of the user within the different rooms of the domestic environment. To constantly control the user's condition, it is fundamental the implementation of an autonomous \textit{Person Following} functionality, exploiting the information retrieved from the perception system. On the other hand, to efficiently accompany the user, the rover must be able to save a series of points of interest, like rooms or specific locations, towards which it can autonomously navigate.
Finally, to facilitate the approach of the user with the remote presence service, a positioning device must be properly designed and deployed on the robotic platform. To this end, it is first necessary to evaluate the workspace requirements. As the application suggests, the tip of the mechanism should reach above common furniture to bring the user interface in a comfortable position. To perform this action, the robot can approach the furniture parking as close as possible to the goal position (Figure \ref{fig:PosDev_MountingStr}{{b},{c}}
) or it can go under the furniture if the cabinetry geometry allows it (Figure \ref{fig:PosDev_MountingStr}{a}). Moreover, the device can be mounted near the closer or on the opposite side of the approached entity. Regarding Figure \ref{fig:PosDev_MountingStr}, the mounting configuration ({Figure \ref{fig:PosDev_MountingStr}}{b}) 
guarantees a better redistribution of the masses to keep the center of mass inside the footprint of the robot, but the device has limited capability at reaching distant points in the longitudinal direction, while the mounting configuration ({Figure \ref{fig:PosDev_MountingStr}}{c}) enables the device to further reach out in that direction, but it moves the center of gravity away from the center of the platform. The workspace related requirements for the positioning device have been chosen considering the following situations:

\begin{itemize}
    \item Dinner table: under motion is possible, no longitudinal displacement from the platform border is required, working height approximately 90--100 cm;
    \item Home bed: under motion is usually not possible, required longitudinal displacement from the platform border of approximately 20 cm, working height approximately 80--90 cm;
    \item Hospital bed: under motion is usually possible, required longitudinal displacement from the platform border of approximately 20 cm, working height approximately 100--110 cm;
    \item Standing person: no longitudinal displacement from the platform border is required, working height approximately 120 cm;
    \item Seated person: required longitudinal displacement from the platform border of approximately 10--20 cm, working height approximately 90--100 cm;
    \item Person on wheelchair: required longitudinal displacement from the platform border of approximately 10--20 cm, working height approximately 80--90 cm;
\end{itemize}

To keep the center of gravity low during the motions, the conceived device needs to be retracted as much as possible. Considering the lightweight requirements, the mounting configuration ({Figure \ref{fig:PosDev_MountingStr}}{b}) seems more suitable for this application because of good weight distribution. The positioning device could also be useful to elevate the RGB camera above most obstacles, to facilitate the tracking of the user in a cluttered environment.

\vspace{-6pt}

Given the set of tasks to be addressed and the application workspace, some specifications can be worked out for both the robotic base platform and the software design.

\begin{itemize}
    \item \textbf{Performance:} as anticipated, the robot should act as a personal assistant. As a consequence, it should be able to follow the user to provide basic assistance. From a mechanical point of view, this implies the need to perform a velocity similar to that of a human walk ($v \approx$ 1--1.5 m/s). The robot should be capable of reaching such cruise velocity in a reasonably short time ($t \approx$ 1--1.5 s): it follows an acceptable maximum acceleration range $a_{MAX}\approx$ 0.7--1.5 m/s$^2$.

    \item \textbf{Dimensions:} the environment where the robot should navigate is designed for human needs. To effectively move in this environment, the assistant should have the same footprint as humans have: maximum encumbrance on the ground approximately of 40 cm $\times$ 60 cm.
    
    \item \textbf{Mobility:} the use case requires the robot to exhibit remarkable mobility to maintain a reduced distance from the user while he is moving within the domestic environment. To such an aim, it turns crucial to provide the mobile platform with full in-plane mobility. Such a feature allows the robot to exhibit velocities in the plane independently from its configuration (orientation).
    
    \item \textbf{Usability:} the identified users’ category suggests that the robot should own an easy-accessible interface to allow efficient interfacing. This feature yields requirements for both software and structural design areas. From the mechanical point of view, the robot layout must allow simple and comfortable access to the interface area. Then, it is interesting to consider the possibility of providing the interface with a proper number of degrees of freedom to make it approach the users’ reach when they are unable to.  
    
    \item \textbf{Computational capability:} given the complexity of the software system, a certain degree of computational resources are needed. In particular, one of the most critical components in this regard is represented by the navigation system: to guide the platform in a cluttered, dynamic environment, it needs to re-plan the optimized path very quickly and react appropriately to very different and potentially dangerous situations. In any case, the final implementation needs to be executed completely on board the platform, without the help of external or remote contributions.
    
    \item \textbf{Modularity:} although the capacities required by the robot and presented before already cover a broad use case, the needs and problems of the domestic environment are constantly evolving. This means that the robot should be able to cope with problems not considered in an early design phase, with the integration of new skills. For this reason, the platform should be designed to be as modular as possible, easily allowing the inclusion or removal of new components, hardware and software. This is necessary also because hardware and software are strictly linked, as an increasing complexity of the application system would require more computational capabilities.
\end{itemize}  

\section{Marvin Modular Approach}
\label{sec:marvin_concept}

\begin{figure}[H]
\centering
\includegraphics[width=13.5 cm]{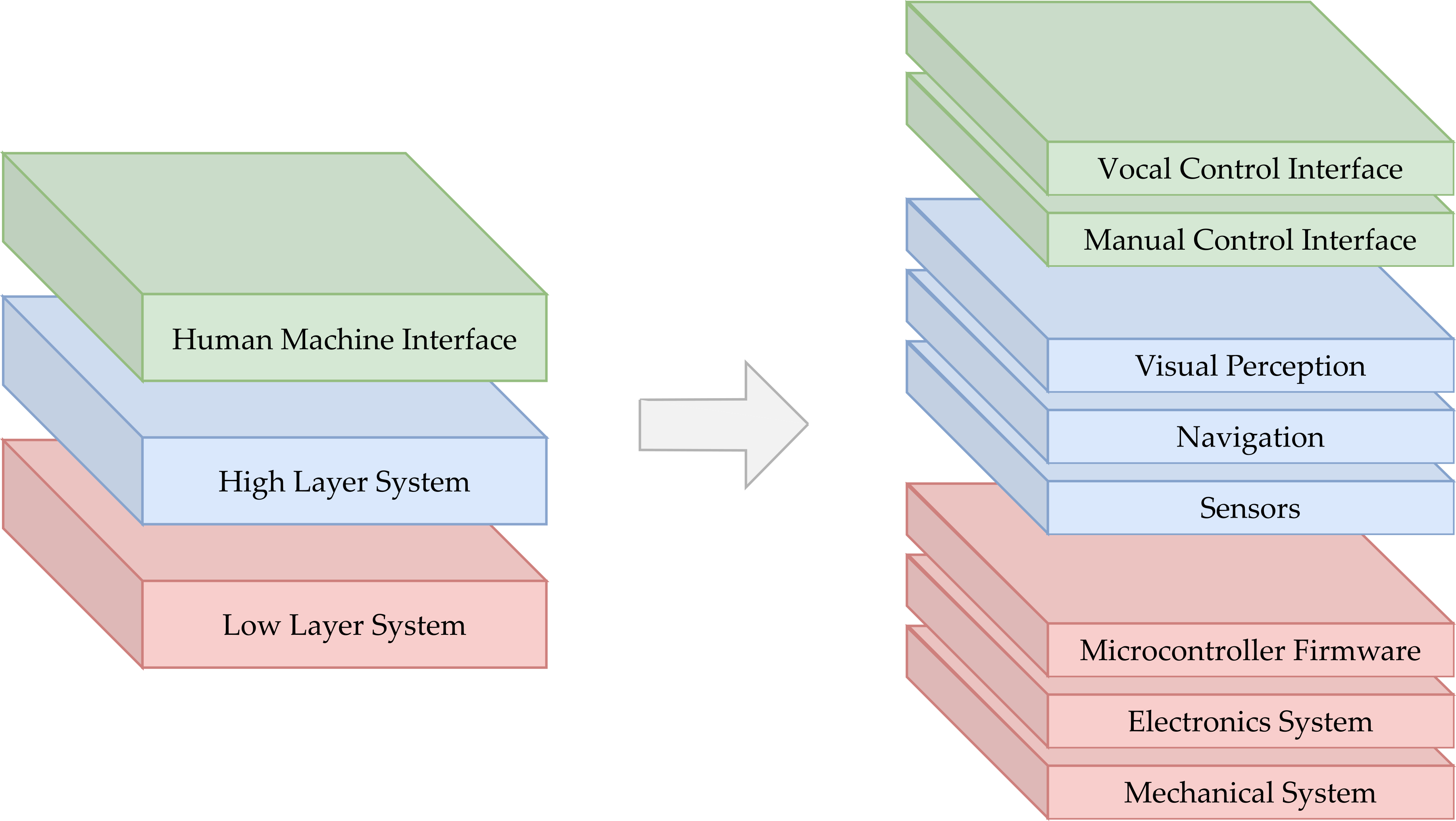}
\caption{Schematic representation of the modular platform architecture. The three main layers of the architecture (\textbf{left}) are decomposed in their respective principal components (\textbf{right}).
\label{fig:architecture}}
\end{figure}

The architecture of the mobile robot has been developed with a modular approach in order to design a robust architecture for both the physical and non-physical constituent parts such as mechanics, electronics, and software. The goal is to obtain a system where small changes, in either the application environment or the implemented features, do not require structural modifications of the system itself. The overall system can be divided into three main layers, as presented in Figure \ref{fig:architecture}:
\begin{itemize}
    \item A \textit{Low Layer System} consists of the mechanical structure, the control electronics, and firmware. This layer is responsible for the actuation and control of the system motion given the desired state of the system which is computed by the Upper Layer System. 
    \item A \textit{Upper Layer System} collects the Upper Layer sensors such as lidar, cameras and remote controller, the autonomous navigation stack, and the visual perception sub-system. This module collects data from the sensors and plans the robot response based on the user's commands interpreted by the Human-Machine Interface and on the current state of the robot. 
    \item A \textit{Human-Machine Interface} consists of a vocal control interface and a manual control interface. The implementation of a custom graphic interface has been discussed but postponed because, even if it enriches the user experience, it does not increase the functionality of the robot which is the main goal of this prototype.
\end{itemize}

The interaction between the different layers is coordinated by predefined communication protocols. In the following sections, the main features of these modules are presented. 

\section{Low Layer System}
\label{sec:low_layer}
As the first design step, it is necessary to identify the typology of a suitable base platform on which to build the entire system. The selected platform should fulfill all requirements as best as possible. This preliminary choice is of fundamental importance, as an evaluation error in this phase could affect the following implementations.
At first, various differential drive platforms are considered for inspiration, like the Robotis TurtleBot3 Waffle \cite{TurtleBot3} and the TurtleBot2 \cite{TurtleBot2}, differential drive robots specially designed taking into account modularity and prototyping. An alternative is the RosBot2 \cite{RosBot2Pro}, a four-wheel differential drive robot.

To overcome the limitations of differential drive locomotion systems and fulfill mobility requirements, an omnidirectional platform is more suitable for the application. Different solutions can be adopted to achieve this level of maneuverability. For example, specially designed wheels \cite{taheri2020OmniMobileRobot}, such as Mecanum wheel \cite{ilon1975,pin1994,salih2006}, Universal wheels \cite{cuevas2019}, Orthogonal wheels \cite{mourioux2006}, Spherical/Ball wheels \cite{tadakuma2007,ferriere1998}, or conventional steerable wheels \cite{ferland2010} can be adopted. In particular, four mecanum wheels and three omniwheels configurations are the most commonly adopted. Reasons for this are the simple control strategy required, omnidirectional mobility with fast response to turn, and simple setup. This improved maneuverability comes with some drawbacks such as discontinuous contact with the ground, a higher sensitivity to floor condition compared to conventional wheels, and payload limitations. Considering the specific environment, ground conditions are quite controlled in indoor applications, even if small obstacles can be found on the floor, while payload limitation is not a problem considering the limited weight of the system. To speed up the prototyping phase, research on commercial solutions has been made. The selected platform, a Nexus 4WD Mecanum robot \cite{Nexus4WD}, is characterized by overall dimensions of \mbox{400 mm $ \times $ 360 mm $ \times $ 100 mm} and a limited mass (5.4 kg), with a passive roll joint between the front wheels and the rear wheels to deal with the presence of four contact points with the ground.

\begin{figure}[H]
\centering
\includegraphics[]{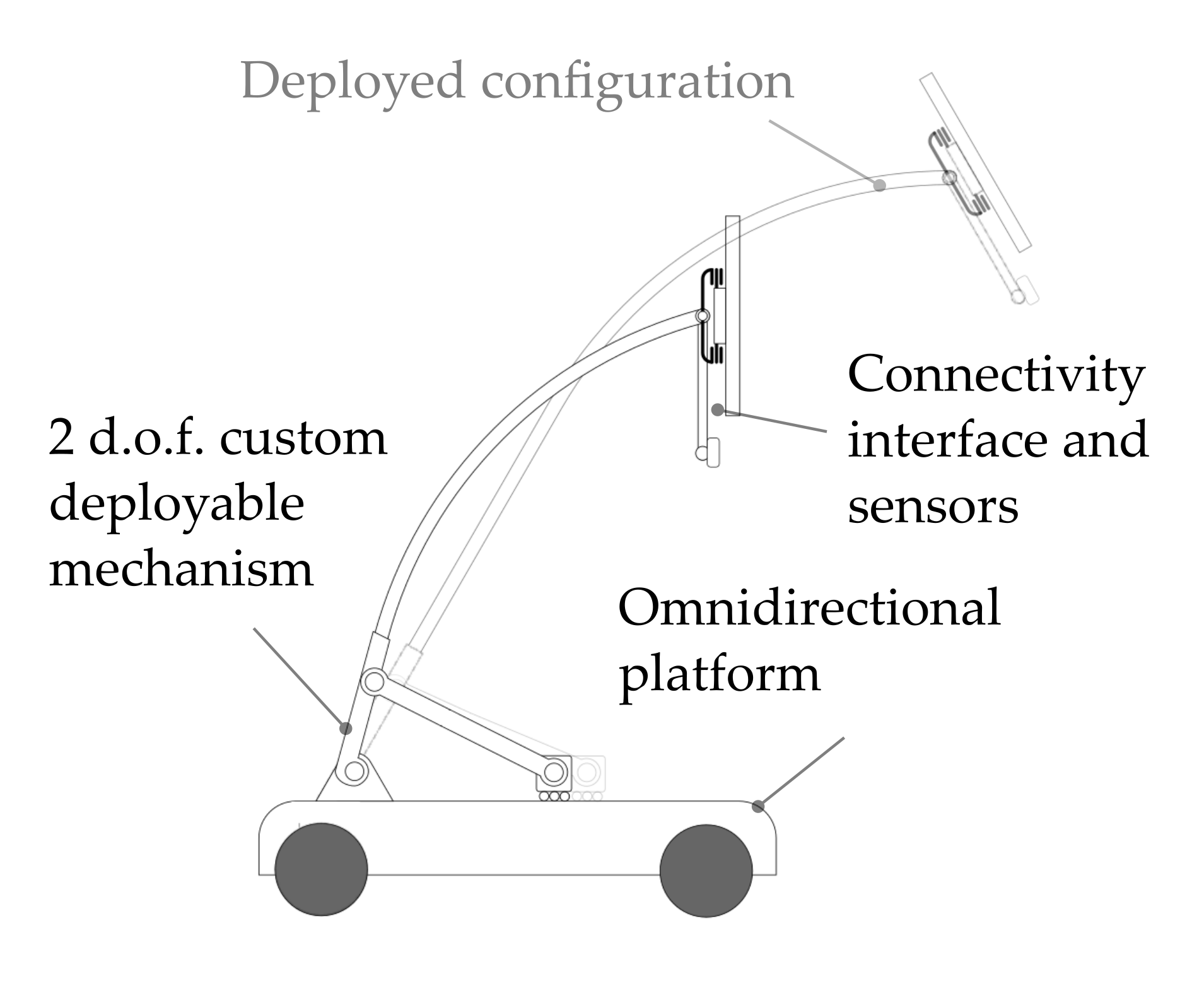}
\caption{Concept representation of the mobile robot.\label{fig:concept}}
\end{figure}

The main peculiarity of the robot, aside from its ability to exhibit full planar mobility, is its capability to deploy its sensors and user interface, exploiting the integrated positioning device (Figure \ref{fig:concept}). Such aspect is crucial for different reasons:
\begin{itemize}
    \item it allows improving the perception of the robot of the external environment improving the range of view of the sensors;
    \item a re-orientable and deployable head enhances the usability of the touch interface for the users, giving a chance also to bedridden or handicapped people to easily interact with the robot.
\end{itemize}\vspace{-6pt}

\begin{figure}[H]
\centering
\includegraphics[width=15 cm]{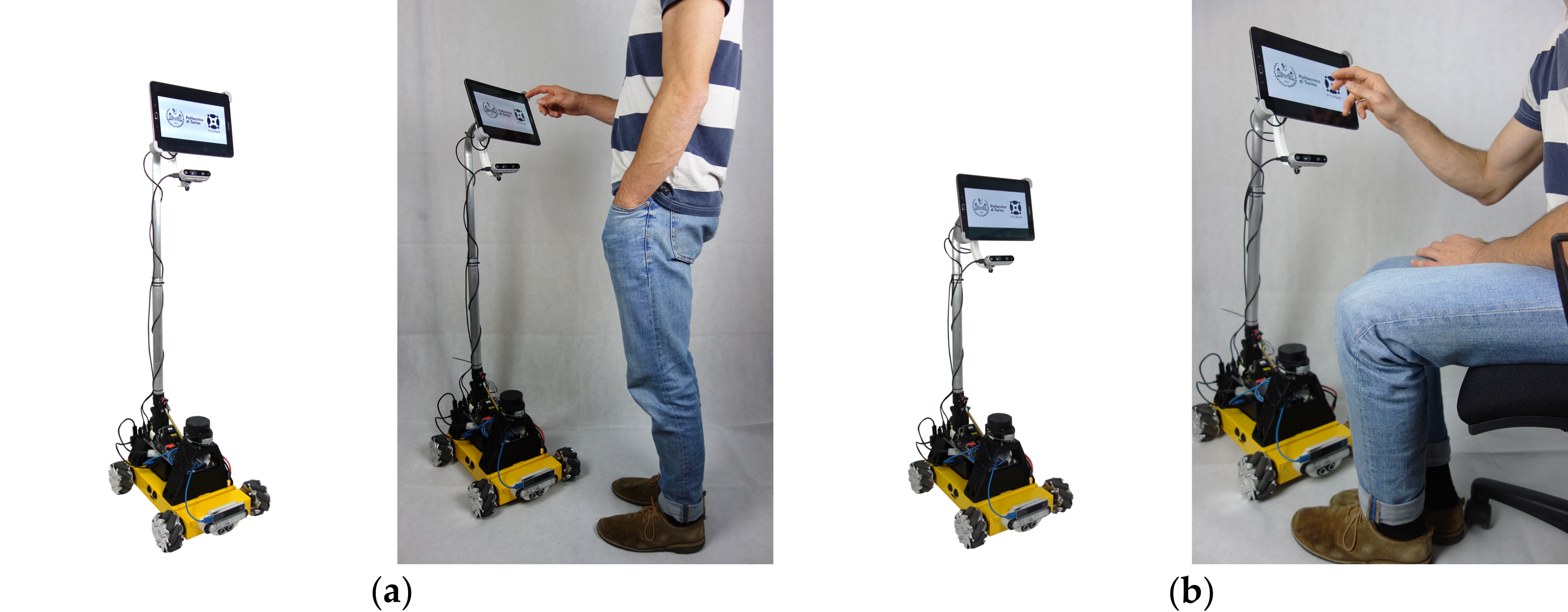}
\caption{Final prototype of the mobile assistive robot in two different working configurations: (\textbf{a}) Deployed configuration for standing usage, HMI height = 1.1 m, mechanism tilting angle $= 0$°, (\textbf{b}) Retracted end angled configuration for better-seated usage, HMI height = 0.80 m, mechanism tilting angle  $= 20$°.}
\label{fig:MARVIN_WorkConf}
\end{figure}

In Figure \ref{fig:MARVIN_WorkConf}, the mobile assistive robot is represented in two configurations: on the left, the telescopic mechanism is deployed for better standing usage, while on the right, the custom mechanism is retracted and inclined forward for better-seated usage. The retracted configuration is also very effective at keeping the center of gravity low during the motion of the robot.

The electronics system firmware is running on the MCU, a {PJRC Teensy 4.1 microcontroller ({\url{https://www.pjrc.com/store/teensy41.html}} (accessed on 1 May 2022)}),
  which is responsible for receiving instructions from the computing unit and acting on the actuators to control the motion of the system.

\section{Upper Layer System}
\label{sec:high_layer}
The Upper Layer System contains all the algorithms for the execution of the various task performed by the robot. Each piece of software dedicated to a specific action of the rover needs to exchange information with the rest of the system to fulfill its assignment properly. To this end, the most widespread solution in literature requires the use of a Middleware \cite{robotics7030047}, an abstraction layer that resides between the operating system and software applications. 
In this project, we decided to adopt the {Robot Operating System 2} (Open Source Robotics Foundation, Inc. ({\url{https://www.openrobotics.org/}} (accessed on \mbox{1 May 2022})))
 (ROS2) \cite{ROS2} due to the variety of compatible algorithms and the very active community supporting it. We preferred it over the original ROS \cite{ROS} as it is more suitable for real-time systems and has access to more advanced applications \cite{ExploringThePerformance,ROS2vsROS1}. ROS2 is based on a Data Distribution Service (DDS) structure, with nodes able to publish and subscribe to different topics.

In our system, all the nodes listen to (or publish on) a specific topic, called \textit{Actions topic}, which contains information about the actual state of the robot and receives requests to perform a new action. This topic is essential to ensure a certain degree of synchronization between all the software components, with consequent robustness of the entire system. 
To control the platform, the user can use two different human-machine interfaces. The first is the Vocal Command (which will be presented in the next chapter), and the second is a wireless gamepad. The latter allows manual control of the platform, as well as the execution of all the tasks. The manual control interface also provides the possibility to send an emergency signal which immediately disables the platform's current action, guaranteeing safety conditions and risk prevention.

\subsection{Sensors}
\begin{figure}[H]
\centering 
\includegraphics[width=17 cm]{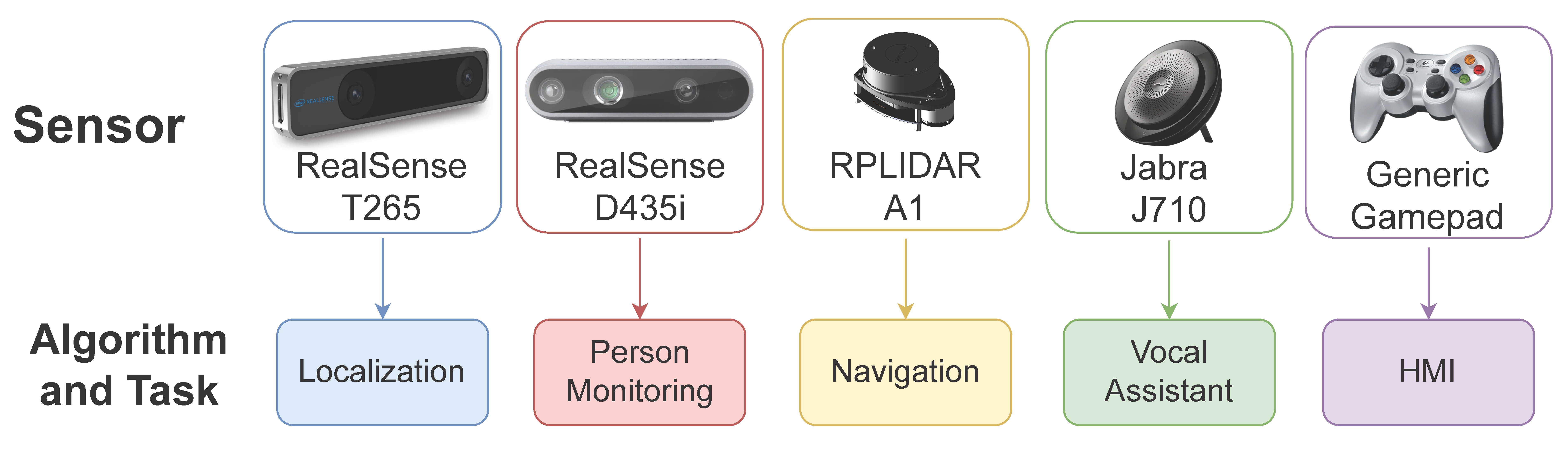}
\caption{Sensors employed on the Marvin robotic platform, associated with the corresponding task they serve.}
\label{fig:Sensors}
\end{figure}

For the robot to effectively work in the domestic environment, a whole series of sensors are required to perceive the surroundings adequately. Alongside classic devices like RGB cameras and Lidar sensors, technology has introduced more powerful tools able to achieve advanced tasks, like self-localization and depth estimation autonomously.
On our robotic platform, some of these state-of-the-art devices are employed (Figure \ref{fig:Sensors}). In particular, the following sensors are used:
\begin{itemize}
    \item {Intel RealSense T265} Tracking Camera\footnote{\url{https://www.intelrealsense.com/tracking-camera-t265/} (accessed on 1 May 2022)} ,
    with VIO technology for self-localization of the platform. It is placed in the front of the rover, to better exploit its capability
    \item {Intel RealSense D435i} Depth Camera\footnote{\url{https://www.intelrealsense.com/depth-camera-d435i/} (accessed on 1 May 2022)}, able to provide color and depth images of the environment. It is mounted on the appropriate support, on the positioning device, which provides a convenient elevated position for the camera
    \item {RPLIDAR A1} \footnote{\url{https://www.slamtec.com/en/Lidar/A1} (accessed on 1 May 2022)}, exploited for its precision in obstacle detection, a fundamental aspect for obstacle avoidance navigation and mapping of the environment. It is mounted on the platform with a specific structure, capable of elevating it above the other components of the robot. In this way, the only blind spot of the sensor is constituted by the rod of the positioning device, which, however, occupies a very limited area and does not compromise the correct functioning of the sensor
    \item {Jabra 710} \footnote{\url{https://www.jabra.com/business/speakerphones/jabra-speak-series/jabra-speak-710##7710-409} (accessed on 1 May 2022)} with a panoramic microphone and speaker. It is particularly useful for voice command. Can be placed on the rover or used wireless from a distance  
    \item Furthermore, a wireless gamepad is employed for manual control operations.
\end{itemize}

\vspace{-6pt}

\subsection{Computational Resources}
With technological advancement, software algorithms have exponentially grown in complexity and computational requirements, leading to the abandonment of limited integrated systems in favor of more powerful hardware. Fortunately, these systems are increasingly widespread and are easy to find, allowing researchers to focus on the development of new applications without worrying about hardware limits.

Our system relies on two fundamental components: the \textit{microcontroller unit} (MCU), which manages the low layer system software and a computing unit that executes all Upper Layer system applications.
The selected microcontroller, a Teensy 4.1, is chosen for its high clock frequency and excellent general performance. It allows communication between the Upper Layer algorithms and the mechanical and electronic system, and vice versa.
On the high-level system, an Intel NUC11TNHv5 is selected as a computing unit, as it represents a good trade-off between high computational power and low energy consumption.
A Coral Edge TPU Accelerator is employed alongside the computing unit to run optimized neural network models without the necessity of a full-size graphics processing unit.

\subsection{Visual Perception for Person Monitoring}
\label{subsec:CV}

Computer vision is a fundamental component of most recent service robotics platforms. In the last decade, Deep Neural Networks (DNN) have largely been demonstrated to be meaningful solutions for a wide variety of visual perception tasks such as real-time object detection \cite{zhao2019object}, semantic segmentation \cite{zhang2018fast} or pose estimation \cite{cao2019openpose}. Robots can exploit the vision of the surrounding environment to extrapolate information and plan their actions accordingly. Nonetheless, visual perception is an extremely effective method for monitoring a person in a domestic scenario. 
We developed a visual perception system for Marvin that contextually detect and track a person from color images. Such information is translated into an effective method for constantly monitoring sudden emergency health conditions of the assisted individual. 
The RealSense D435i API is used to retrieve aligned color and depth images. 

The monitoring task is carried out through a double-step computing pipeline. Firstly, the person-detection is obtained with PoseNet \cite{PoseNet}, a lightweight neural network able to detect humans in images and videos. As output, it gives 17 key joints (like elbows, shoulders, or feet) of each person present in the scene.  At this point, a second simple convolutional neural network (CNN) receives the key points to classify the pose of the person as standing, sitting, or laying. As already explained, a persisting laying condition can automatically activate an emergency call to an external agent (a relative or a healthcare operator). A custom labeled dataset of images has been collected in a house environment to train the CNN for the pose classification. The total number of images used for this dataset is 25,009. The images are divided into three classes: standing, sitting, and lying, containing 7849, 11,400, and 5760 images, respectively. The classification model reaches an accuracy of almost \(99\%\) on the test set, obtained retaining the \(20\%\) of the original dataset. The performance of the model is definitely high, probably due to the common background scene of the collected images. A randomized background with scenes of diverse domestic environments may allow for a more challenging testing condition, leading also to improving the generalization performance of the model. Besides the accuracy results, an important remark is that the whole pose estimation algorithm has been drastically optimized to guarantee a real-time monitoring system. Moreover, the lightweight model runs on the Google Coral Edge TPU device for a faster inference: the CNN is able to run at 30 frame-per-second (FPS), which is the maximum frame rate allowed with the Realsense D435i camera.

\begin{figure}[H]
\centering
\includegraphics[width=13 cm]{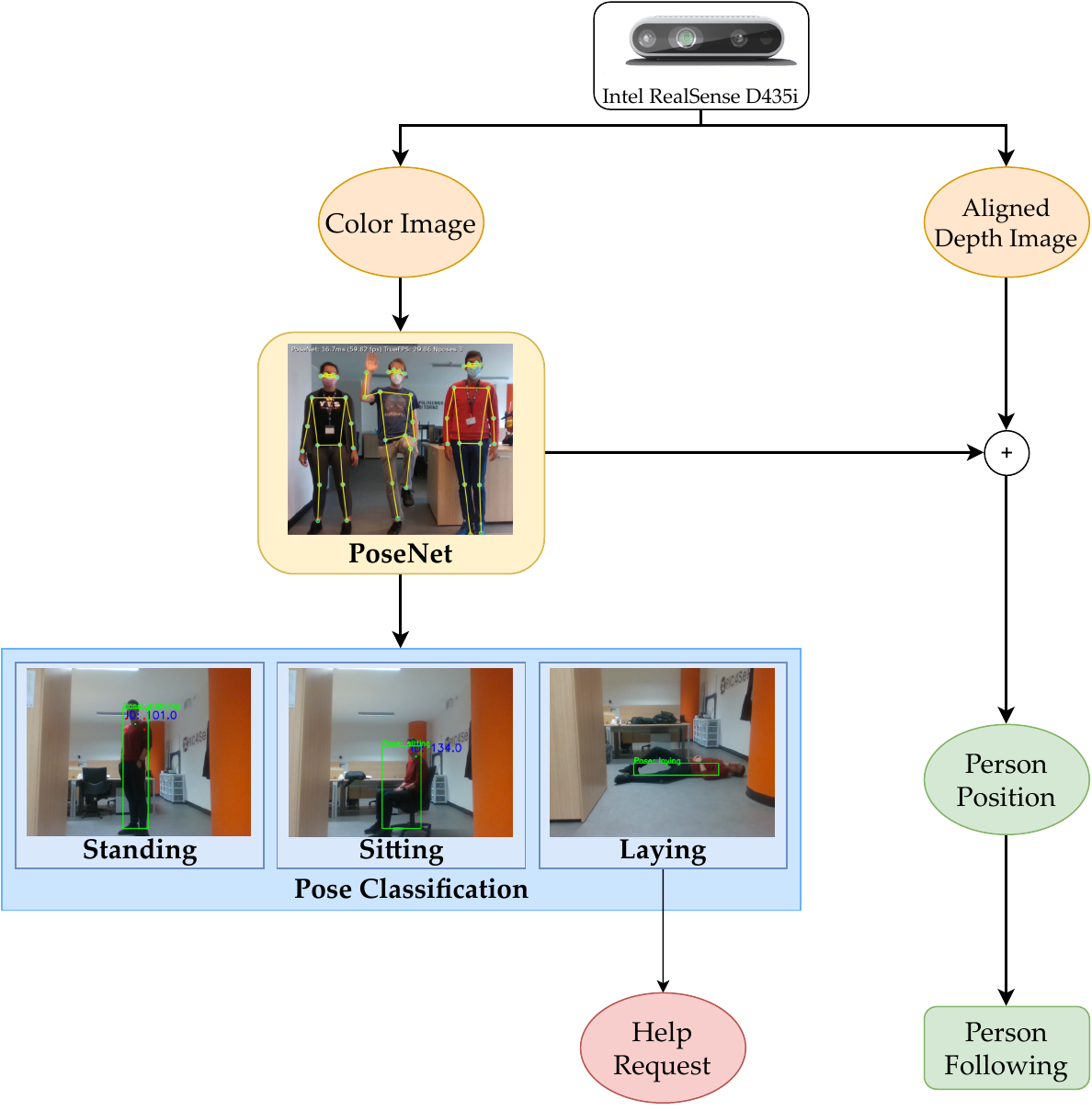}
\caption{{Representation}
 diagram of the person identification system: the estimated pose of the person is continuously classified as standing, sitting, or laying, generating a help emergency request if necessary. Moreover, it is used to extract the dynamic goal coordinates for the person following task.}
\label{fig:person_following_scheme}
\end{figure}

Moreover, as shown in Figure \ref{fig:person_following_scheme}, the key points predicted by PoseNet for the detected person can be exploited for a different assistive task: the person following. Indeed, once a person is recognized within the color image, it is possible to derive the coordinates of such person with respect to the robot from the aligned depth image at any instant. This constitutes an effective method to generate a dynamic goal, corresponding to the position of the user, to be reached by the robot. The person-following navigation system is fed with such information and allows the robot to follow the user around the house. However, more than a single person is usually present in a family house, dramatically increasing the difficulty of an automatic system to recognize the person to follow. For this reason, we combine in our visual perception software a filter called Sort \cite{Sort}, which is used within the same node to track the people recognized. In particular, it assigns an ID to each person in the image and tracks them during their motion. The person with the lowest ID is chosen to be followed: the robot focuses always on a single person and the computational complexity of the task is considerably reduced. Nonetheless, a further improvement of the person-following task can be achieved by adopting a neural network for person re-identification, allowing the robot to discard undesired detected persons and reduce the interference with its monitoring activity.

\subsection{Navigation and Mapping System}
\label{subsecsec:navigation}
In any navigation system, the primary necessity consists in localizing the robot within the operating scenario. To achieve this, the localization node exploits the Intel RealSense API to communicate with the T265 camera and get the pose of the rover at any time instant, through visual-inertial odometry technology \cite{debeunne2020review}.
The navigation system is based on the Navigation2 stack \cite{macenski2020marathon}, which has been highly modified to suit the needs of the platform. Details on the entire development and optimization process of the navigation behavioral tree are out of the scope of this paper. When the rover is asked to reach a specific goal or to follow the use, the navigation system retrieves the pose of the rover and exploits the 2D LiDAR points to perceive surrounding obstacles and create a local cost map. From such a cost map, the navigation apparatus plans an optimal path for the platform and guide it towards the desired destination.
Similarly, the mapping system, based on Slam Toolbox \cite{Slam_Toolbox}, uses the pose of the rover and the laser scan to generate a grid map of the environment. Although the navigation system perfectly adapts to mapless circumstances, the generated map of the domestic environment can be saved by the robot.

\section{Vocal Human--Robot Interface}
\label{sec:HRInterface}

\begin{figure}[H]
\centering 
\includegraphics[width=18 cm]{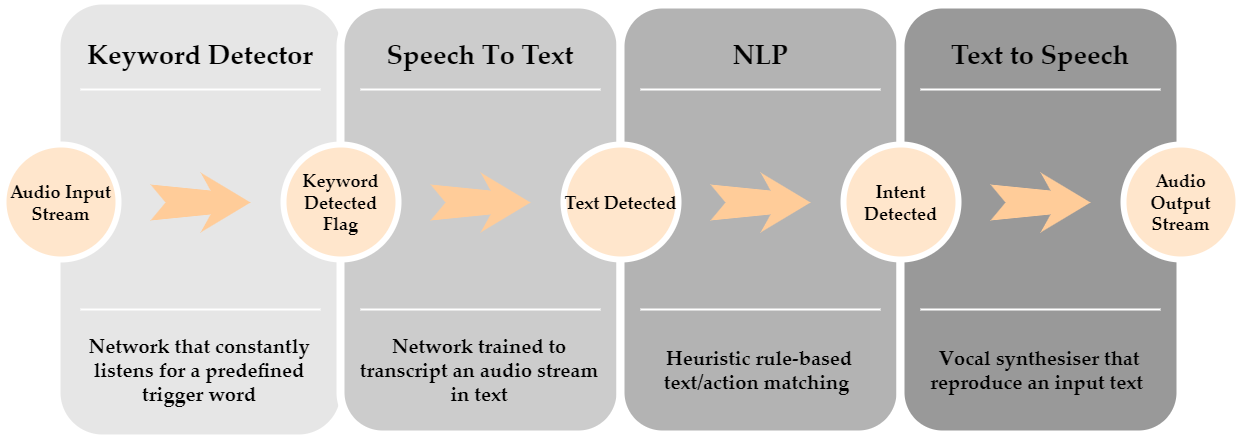}
\caption{Overview of PIC4Speech vocal assistant architecture. The scheme describes the successive cascade activation of the different components of the vocal assistant pipeline.}
\label{fig:vocal_scheme}
\end{figure}

The principal user's communication interface with the platform is represented by an offline vocal assistant. We build our vocal assistant system, called PIC4Speech, exploiting the combination of state-of-the-art Deep Neural Networks (DNN) for speech-to-text translation and a simple rule-based model for Natural Language Processing (NLP) taken from literature with the aim of minimizing the computational cost of the pipeline and preserving a flexible interaction. The overall structure of the system is inspired by the most notable products: Siri, Alexa, and Google Assistant. It exploits a cascade of models that are progressively activated when the previous one is enabled. In Figure \ref{fig:vocal_scheme}, an overview of the overall architecture of the PIC4Speech vocal assistant is represented, showing the subsequent activation of each block. 
Although PIC4Speech is designed as an offline vocal assistant, its usage in this primitive version is mainly devoted to allowing the user to give commands to the robot vocally and not to hold a complete conversation. In particular, the system aims at matching a vocal instruction expressed by the user to the corresponding required task to successively start the correct control process by publishing a ROS message on the Actions topic.

The PIC4Speech operative chain can be summarized as follows:
\begin{enumerate}
    \item The first component is the \textbf{keyword detector} \cite{berg2021keyword}, which constantly monitors the input audio stream in search of the specific triggering command. In this specific case, that word is the name of the robot: ``Marvin.''
    \item Once the trigger word is detected, a second model performs a \textbf{speech-to-text} operation. We exploited the Vosk offline speech recognition API \cite{andreev2021speech} for this block which gives the flexibility to switch language and has ample community support. It continuously analyzes the input audio stream until the volume is below a certain threshold and performs the transcription.
    \item The transcribed text is subsequently passed to a \textbf{natural language processing (NLP)} algorithm that matches the input with a certain number of predefined intents. The recognized robot action is therefore published on the Actions topic of the ROS framework.
    \item The response of the vocal assistant is also given to the user with a \textbf{text-to-speech} process. Each OS comes with a default vocal synthesizer that can directly access the speakers. 
\end{enumerate}

More in detail, the keyword detection is performed with a DNN based on a Vision Transformer that constantly listens to the audio stream, looking for the target command. First, the mel-scale spectrogram is extracted from each sample of the input stream. These features are treated as visual information and, therefore, they are processed with a Vision Transformer \cite{dosovitskiy2020image}, a state-of-the-art model for image classification. 
We re-trained the keyword detector from scratch on the Speech Commands dataset \cite{warden2018speech}, constituted by 1-second-long audio samples from 36 classes: 35 standard keywords plus a silence/noise class. The re-train model achieved a test accuracy of \(97\%\) over the different classes on the 11,005 test samples of the Speech Commands dataset. Specifically, for the current target class ‘Marvin’, we get the results reported in Table \ref{tab:Keyword_results}, evaluating the performance with standard classification metrics: $Precision=TP/(TP+FP)$, $Recall=TP/(TP+FN)$ and $F1 score=2\cdot(recall\cdot precision)/(recall+precision)$.

Thanks to the multi-class approach, the keyword can be changed at run-time.
Being constantly active, it is of primary importance that this network consumes less energy as possible, but at the same time, it is capable of maintaining a good compromise between false positive and false negative detections. At the same time, the network should deal with different sound environments and noise levels. Further improvements to the keyword detector can be reached by augmenting the training set with newly generated samples to increase the robustness of the model in crowded, noisy environments.
At the moment, a simple rule-based matching mechanism based on keywords is used for the NLP stage. Although it represents a simple approach, a good level of flexibility is guaranteed by the actual solution as the user can introduce new actions for the robotic platform associated with several indicative sentences. Future works may involve the investigation of suitable DNN models for the NLP stage of PIC4Speech, with the aim of semantically matching the encoded query text and the robot actions.

\begin{table}[ht]
\centering
\caption{Classification results of the target keyword `Marvin' on the 11,005 test samples of the Speech Commands dataset. Standard classification metrics are used for the evaluation:  $Precision=TP/(TP+FP)$, $Recall=TP/(TP+FN)$ and $F1 score=2\cdot(Recall\cdot Precision)/(Recall+Precision)$.}
\label{tab:Keyword_results}
\newcolumntype{C}{>{\centering\arraybackslash}X}
\begin{tabularx}{0.8\textwidth}{CC}
\toprule
\makecell{\textbf{Keyword Detector Classification Metrics}} & \textbf{Results} \\ 
\midrule
\makecell{True Positives ({$TP$})
} & $189$ \\ 
\midrule
\makecell{True Negatives ({$TN$})} & 10,806 \\
\midrule
\makecell{False Positives ({$FP$})} & $4$ \\ 
\midrule
\makecell{False Negatives ({$TN$})} & $6$ \\
\midrule
\makecell{{$F1 Score$}} & $0.9742$  \\
\midrule
\makecell{{$Precision$}} & $0.9793$ \\
\midrule
\makecell{{$Recall$}} & $0.9692$  \\
\bottomrule
\end{tabularx}
\end{table}

Different from commercial vocal assistants, which require a stable internet connection, PIC4Speech works completely offline, running uniquely on the hardware resources of the platform. This competitive advantage derived from the choice of lightweight models in the algorithms pipeline prevents Marvin from exposing the visual data of the domestic environment to internet-derived risks.
Diversely, companies prefer to offer a server-client paradigm with the assistant algorithms running on the cloud. That solution presents some computational advantages and enables the connectivity of the platform with the whole house. Furthermore, it dramatically facilitates data collection. This solution may raise privacy issues related to the collection of visual and vocal data collected in domestic private environments currently used on the robot systems. In addition, the dependency on a stable internet connection may weaken the system performance in terms of response time and power consumption. For all those reasons, an offline solution should largely fit most service robotics applications requiring vocal control.

Moreover, it is worth noting that a help request is constantly visually checked by the pose classification node but it can also be called directly through vocal command. In this case, the platform will ask the user for confirmation, avoiding any accidental activation. If confirmed, without any reply within ten seconds, the help request is sent. Otherwise, the platform return to its regular operation state.
Further development of the PIC4Speech vocal assistant could see the substitution of the last block for text-to-speech conversion with an additional lightweight neural network, also providing the possibility to choose a more comfortable synthetic voice closer to a real human one.

\section{Experimental Demo}
\label{sec:demo}
\begin{figure}[ht]
\centering
\includegraphics[width=15 cm]{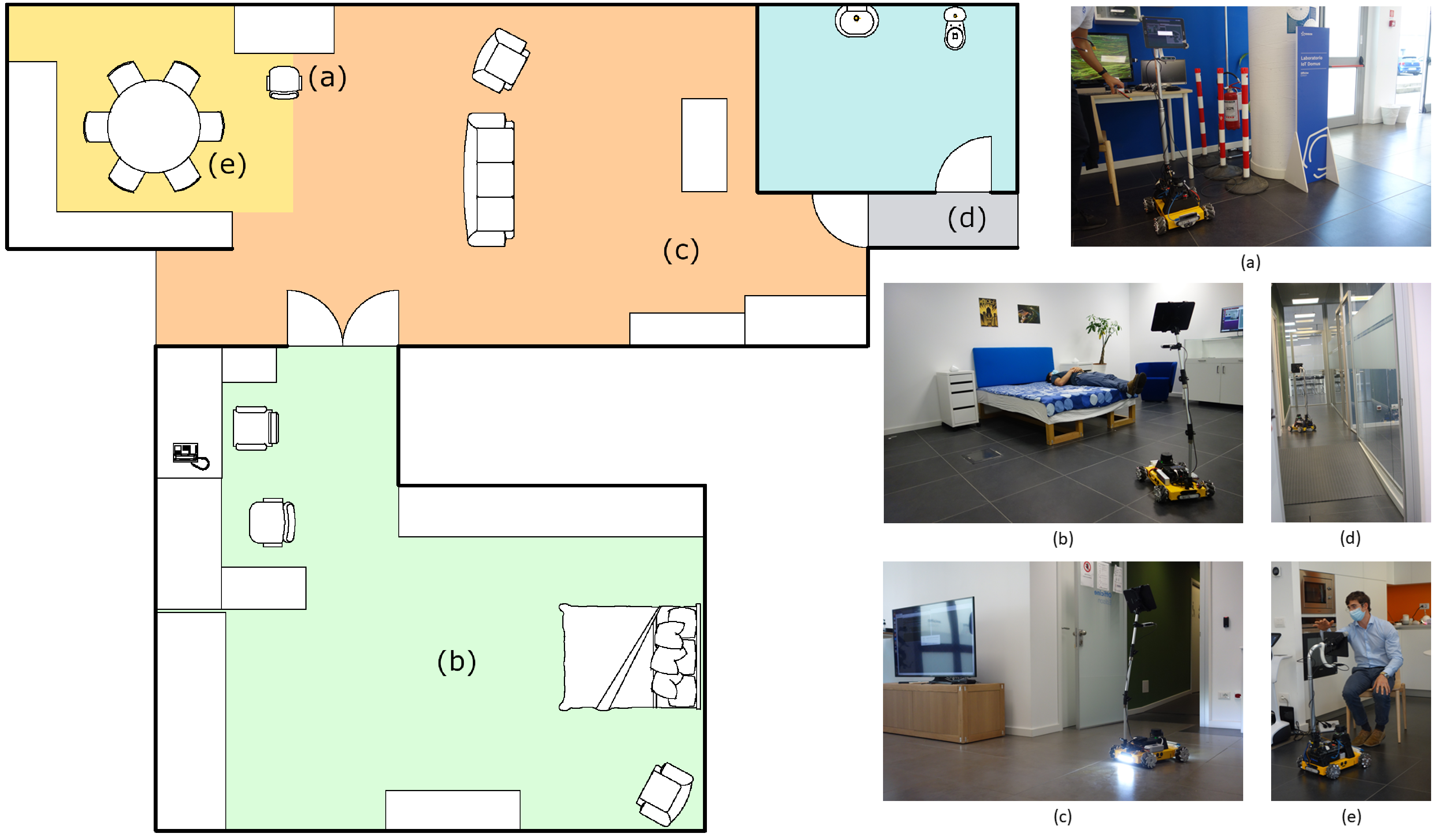}
\caption{Simplified map of the Domus area at Officine Edison, Milan, with the four rooms, kitchen, living room, bedroom, and bathroom, respectively in yellow, orange, green, and blue. Letters on the map indicate the various goals saved in the environment, associated with the corresponding image: (\textbf{a}) starting point, (\textbf{b}) bedroom keypoint, with user's pose recognition, (\textbf{c}) living room keypoint, with lights turned on, ready for night assistant task (\textbf{d}) bathroom keypoint, (\textbf{e}) kitchen keypoint, with demonstration of the positioning device capabilities.}
\label{fig:Domus}
\end{figure}

We conducted a qualitative demonstration of the platform's capabilities at Officine Edison, Milan, during an institutional presentation specifically organized to test and show Marvin and validate the outcome of its prototyping process. The demonstration took place in an area called Domus (Figure \ref{fig:Domus}), which simulates a real domestic environment made up of a kitchen, bedroom, living room, and bathroom. Like a normal house, the Domus features different obstacles of various heights and dimensions, and rooms are separated by regular size doors. 

In the setup phase, Marvin was guided in each of the different rooms and their relative positions were saved with respect to the starting point, where a docking station for recharging could eventually be placed. Moreover, a telephonic number was memorized for the emergency call task.
In the demonstration, all the functionalities of the robot were tested, and a qualitative analysis was conducted. From the starting point ({Figure}
 \ref{fig:Domus}a), where a brief introduction was given to attendees, the rover was asked to autonomously reach the bedroom waypoint, passing through the double-leaf door. Here, the user monitor function was demonstrated, showing how Marvin was able to correctly classify the pose of the visualized user, standing, sitting on an armchair, or laying in the bed ({Figure} \ref{fig:Domus}b). In addition, it was also demonstrated how, after a request from the user, the system was capable of connecting with the pre-configured telephone to call the emergency number.
Then, the rover was asked to follow the user from the bedroom to the living room \mbox{({Figure} \ref{fig:Domus}c)}. There, the user asked Marvin to save a new semantic waypoint, to demonstrate how the various room destinations within the environment can be saved and used by the robot for autonomous navigation tasks. Visual perception algorithms, namely person detection and pose estimation and classification, have been visually validated publicly showing on the television screen what the robot ``saw'' through cameras. The crowd of attendees received an in-depth explanation of how the various algorithms of person and pose estimation work while looking at the resulting predictions in real-time. Later, the user activated the night assistance task by asking the rover to accompany him to the bathroom, causing the robot to turn on the on board lights ({Figure} \ref{fig:Domus}d).
Finally, Marvin was asked to reach the kitchen and to adjust the inclination and height of the positioning device to adapt to the user, sitting on the chair, so that the mounted tablet could be more easily accessed and operated \mbox{({Figure} \ref{fig:Domus}e).}

\begin{table}[hb]
\centering
\caption{The three service functions (left) provided by the Marvin robot in the context of home assistance are associated with the robotic tasks (right).}
\label{tab:Services&Tasks}
\newcolumntype{C}{>{\centering\arraybackslash}X}
\begin{tabularx}{\textwidth}{CCC}
\toprule
\makecell{\textbf{Service Function}} & \makecell{\textbf{Robotic Tasks}} & \makecell{\textbf{Demo Execution}}\\ 
\midrule
User Monitoring\vspace{-10pt}
 & \makecell{Person Following\\Pose Classification\\Emergency Call} & \makecell{(b) $\implies$ (c)\\ always running \\ (b)}  \\ 
\midrule
Night Assistance\vspace{-5pt} & \makecell{Autonomous Navigation\\Lighting Control} & \makecell{(c) $\implies$ (d)}\vspace{-5pt} \\
\midrule
Remote Presence \& Connectivity\vspace{-5pt} & \makecell{Positioning Device Control\\Autonomous Navigation} & \makecell{(e) \\ (a) $\implies$ (b)} \\
\bottomrule
\end{tabularx}
\end{table}

During the demonstration, we collected some observations about the performance of the system. In particular, we focused our attention on the success of the three proposed service functions, determined by the fulfillment of the required robotic tasks, as reported in Table \ref{tab:Services&Tasks}. Thanks to its mobility and the elevated position of the camera, mounted on the positioning device, the rover managed to efficiently track and follow the person, albeit the several obstacles with diverse heights. This allowed to continuously monitor the user, classifying their pose in any instant, and calling for help during potentially dangerous situations. Moreover, the rover showed no difficulties in autonomously navigating between the various rooms, avoiding static and dynamic obstacles arranged in different configurations, and accompanying the user to the desired destination. In particular, the robot was also able to manage its handling upon less conventional surfaces, such as the skirting board of fire doors and a small ramp placed in the corridor before the bathroom. Finally, the positioning device's flexibility, in conjunction with the platform maneuverability, guaranteed the user an easy operation of the tablet, standing, sitting, or lying down.

Further analyses were conducted to validate the vocal assistant. The vocal command was tested in three different scenarios. The first one took place in an environment characterized by good acoustic conditions and without any relevant noise apart from the tester's voice giving commands. The second test scenario is located in the same environment described before, but with the addition of background noise caused by a group of people talking. The third test scenario was realized in a silent environment presenting not-optimal acoustic conditions, like echo and reverb. 
To evaluate the performance of the vocal assistant, twenty test commands were given for each scenario and six parameters were taken into consideration. Of these, the results proved only three are scenario dependent:
\begin{itemize}
    \item Keyword detector success frequency, indicate how many times the Keyword detector is triggered when the trigger word is pronounced 
    \item Keyword detector accuracy indicate the average confidence over the prediction of the trigger word
    \item Command understanding frequency, indicate how many times the command given after the trigger is correctly understood by the speech-to-text model
\end{itemize}

The other three parameters returned no significant differences among the various scenarios:

\begin{itemize}
    \item Trigger delay, indicate the average time delay between the pronunciation of the trigger word and the actual trigger
    \item Command delay, indicate the average time delay between the given command and the received feedback from the system 
    \item Help request delay, indicate the average time delay between the help request command and the starting of the call towards the emergency number
\end{itemize}

The final results can be consulted in Table \ref{tab:DetectorPerformances}.

\begin{table}[ht]
\caption{Performances of vocal assistant PIC4Speech during the experimental demo.}
\label{tab:DetectorPerformances}

		\newcolumntype{C}{>{\centering\arraybackslash}X}
		\begin{tabularx}{\textwidth}{CCCC}
\toprule
& \makecell{\textbf{Muffled, Silent}\\\textbf{Environment}} & \makecell{\textbf{Muffled Environment}\\\textbf{with People Talking}} & \makecell{\textbf{Environment with Echo}\\\textbf{and Reverb}}\\ 
\midrule
\makecell{\textbf{{Keyword detector}
}\\\textbf{Success rate}} & $100.00\%$\vspace{-7pt}
 & $95.00\%$\vspace{-7pt} & $95.00\%$\vspace{-7pt} \\ 
\midrule
\makecell{\textbf{Keyword detector}\\\textbf{Accuracy}} & $91.14\%$\vspace{-7pt} & $88.24\%$\vspace{-7pt} & $89.04\%$\vspace{-7pt} \\
\midrule
\makecell{\textbf{Command understanding}\\\textbf{Frequency}} & $100.00\%$\vspace{-7pt} & $84.21\%$\vspace{-7pt} & $94.74\%$\vspace{-7pt} \\
\midrule
\makecell{\textbf{Trigger delay}} & & <0.5 s & \\
\midrule
\makecell{\textbf{Command delay}} & & 1.45 s & \\
\midrule
\makecell{\textbf{Help request delay}} & & <0.5 s & \\
\bottomrule
\end{tabularx}
\end{table}

\section{Conclusions and Future Works}
In the era of automatic machines, technology is progressively reshaping the domestic environment as we know it. In particular, service robotics is recalling an ever-growing interest of markets, industries, and researchers. Their exploitation in the caregiving sector could relieve the pressure on assistive operators, providing basic assistance which does not require particular dexterity or adaptation capability. In this scenario, we developed Marvin: a modular assistive mobile robot for autonomous applications in the field of home assistance. In this work, Marvin has been initially presented as a robotic assistant solution tailored for the practical use case of monitoring elderly and reduced-mobility subjects in their domestic environment, although its applicability can be easily extended to the alternative person monitoring scenarios in indoor environments. Hence, this paper aims to fully describe Marvin, a four mecanum-wheel robot provided with a custom positioning device for the human-machine interface and state-of-the-art Artificial Intelligence methods for perception and vocal control. The robot has been fully prototyped and qualitatively tested in a domestic-like environment and it proved to be successful in the execution of the target tasks. 

Future works will firstly try to enrich the experimentation by providing more task-specific experimental results and subsequently extend the applicability of Marvin to unseen service robotics functions. More in detail, a great focus will be devoted to the application of Marvin to person-centered autonomous navigation tasks. A secondary future direction deals with the upgrade of the human--robot interface of the robot, enhancing its proactive behavior in social domestic environments and its awareness of the context through more sophisticated visual techniques. Furthermore, the combination of vocal and visual inputs can help the robot contextualize its actions better, resulting in higher precision in tasks execution.

\section*{Acknowledgments}
The work presented in this paper has born from the collaboration between the PIC4SeR Centre for Service Robotics at Politecnico di Torino and Edison S.p.A. In~particular, we sincerely thank Riccardo Silvestri and Stefano Ginocchio, as~well as the entire team from Officine Edison Milano that fruitfully contributed to the funding and conceptualization of Marvin, and~supervised the whole design process. We demonstrated Marvin’s capabilities in the Smart Home facility at Officine Edison Milano, simulating a real-case domestic assistance scenario, showing how Marvin successfully fulfill the tasks requirements identified in the design~process.

\section*{Funding}
This research was developed by a collaboration between EDISON Spa, grant number 06722600019 and the interdepartmental research group PIC4SeR of Politecnico di Torino, Italy.

\section*{Authors contribution}
Conceptualization, A.E., M.M., L.T., D.G., M.C., and G.Q.; methodology, A.E., M.M., L.T., D.G., M.C., and G.Q.; software, A.E. and M.M.; validation, A.E., M.M., and L.T.; formal analysis, A.E., M.M., and L.T.; investigation, A.E., M.M., and L.T.; writing---original draft preparation, L.T., M.M., and A.E.; writing---review and editing, L.T., M.M., and A.E.; visualization, L.T., M.M., and A.E.; supervision, D.G., M.C., and G.Q.; project administration, D.G., M.C., and G.Q. All authors have read and agreed to the published version of the~manuscript.

\bibliographystyle{unsrt}  
\bibliography{biblio}  

\begin{thebibliography}{10}

\bibitem{UnitedNations}
United Nations.
\newblock Shifting demographics.

\bibitem{vercelli2018robots}
Alessandro Vercelli, Innocenzo Rainero, Ludovico Ciferri, Marina Boido, and
  Fabrizio Pirri.
\newblock Robots in elderly care.
\newblock {\em DigitCult-Scientific Journal on Digital Cultures}, 2(2):37--50,
  2018.

\bibitem{abdi2018scoping}
Jordan Abdi, Ahmed Al-Hindawi, Tiffany Ng, and Marcela~P Vizcaychipi.
\newblock Scoping review on the use of socially assistive robot technology in
  elderly care.
\newblock {\em BMJ open}, 8(2):e018815, 2018.

\bibitem{gouaillier2009mechatronic}
David Gouaillier, Vincent Hugel, Pierre Blazevic, Chris Kilner, J{\'e}r{\^o}me
  Monceaux, Pascal Lafourcade, Brice Marnier, Julien Serre, and Bruno
  Maisonnier.
\newblock Mechatronic design of nao humanoid.
\newblock In {\em 2009 IEEE International Conference on Robotics and
  Automation}, pages 769--774. IEEE, 2009.

\bibitem{fujita2001aibo}
Masahiro Fujita.
\newblock Aibo: Toward the era of digital creatures.
\newblock {\em The International Journal of Robotics Research},
  20(10):781--794, 2001.

\bibitem{vsabanovic2013paro}
Selma {\v{S}}abanovi{\'c}, Casey~C Bennett, Wan-Ling Chang, and Lesa Huber.
\newblock Paro robot affects diverse interaction modalities in group sensory
  therapy for older adults with dementia.
\newblock In {\em 2013 IEEE 13th international conference on rehabilitation
  robotics (ICORR)}, pages 1--6. IEEE, 2013.

\bibitem{gongora2019social}
Susel G{\'o}ngora~Alonso, Sofiane Hamrioui, Isabel de~la Torre~D{\'\i}ez,
  Eduardo Motta~Cruz, Miguel L{\'o}pez-Coronado, and Manuel Franco.
\newblock Social robots for people with aging and dementia: a systematic review
  of literature.
\newblock {\em Telemedicine and e-Health}, 25(7):533--540, 2019.

\bibitem{gasteiger2021friends}
Norina Gasteiger, Kate Loveys, Mikaela Law, and Elizabeth Broadbent.
\newblock Friends from the future: A scoping review of research into robots and
  computer agents to combat loneliness in older people.
\newblock {\em Clinical interventions in aging}, 16:941, 2021.

\bibitem{8448739}
Akihito Yatsuda, Toshiyuki Haramaki, and Hiroaki Nishino.
\newblock A study on robot motions inducing awareness for elderly care.
\newblock In {\em 2018 IEEE International Conference on Consumer
  Electronics-Taiwan (ICCE-TW)}, pages 1--2, 2018.

\bibitem{mundher2014real}
Zaid~A Mundher and Jiaofei Zhong.
\newblock A real-time fall detection system in elderly care using mobile robot
  and kinect sensor.
\newblock {\em International Journal of Materials, Mechanics and
  Manufacturing}, 2(2):133--138, 2014.

\bibitem{saini2021sensors}
Jagriti Saini, Maitreyee Dutta, and Goncalo Marques.
\newblock Sensors for indoor air quality monitoring and assessment through
  internet of things: a systematic review.
\newblock {\em Environmental Monitoring and Assessment}, 193(2):1--32, 2021.

\bibitem{mocrii2018iot}
Dragos Mocrii, Yuxiang Chen, and Petr Musilek.
\newblock Iot-based smart homes: A review of system architecture, software,
  communications, privacy and security.
\newblock {\em Internet of Things}, 1:81--98, 2018.

\bibitem{marques2019air}
Gon{\c{c}}alo Marques, Ivan~Miguel Pires, Nuno Miranda, and Rui Pitarma.
\newblock Air quality monitoring using assistive robots for ambient assisted
  living and enhanced living environments through internet of things.
\newblock {\em Electronics}, 8(12):1375, 2019.

\bibitem{doroftei2007omnidirectional}
Ioan Doroftei, Victor Grosu, and Veaceslav Spinu.
\newblock {\em Omnidirectional mobile robot-design and implementation}.
\newblock INTECH Open Access Publisher, 2007.

\bibitem{al2018embedded}
Md~Abdullah Al~Mamun, Mohammad~Tariq Nasir, and Ahmad Khayyat.
\newblock Embedded system for motion control of an omnidirectional mobile
  robot.
\newblock {\em IEEE Access}, 6:6722--6739, 2018.

\bibitem{costa2016localization}
Paulo~Jos{\'e} Costa, Nuno Moreira, Daniel Campos, Jos{\'e} Gon{\c{c}}alves,
  Jos{\'e} Lima, and Pedro~Lu{\'\i}s Costa.
\newblock Localization and navigation of an omnidirectional mobile robot: the
  robot@ factory case study.
\newblock {\em IEEE Revista Iberoamericana de Tecnologias del Aprendizaje},
  11(1):1--9, 2016.

\bibitem{qian2017design}
Jun Qian, Bin Zi, Daoming Wang, Yangang Ma, and Dan Zhang.
\newblock The design and development of an omni-directional mobile robot
  oriented to an intelligent manufacturing system.
\newblock {\em Sensors}, 17(9):2073, 2017.

\bibitem{jibo}
{Jibo Robot Website}.
\newblock \url{https://jibo.com/}, 2017.

\bibitem{fischinger2016hobbit}
David Fischinger, Peter Einramhof, Konstantinos Papoutsakis, Walter Wohlkinger,
  Peter Mayer, Paul Panek, Stefan Hofmann, Tobias Koertner, Astrid Weiss,
  Antonis Argyros, et~al.
\newblock Hobbit, a care robot supporting independent living at home: First
  prototype and lessons learned.
\newblock {\em Robotics and Autonomous Systems}, 75:60--78, 2016.

\bibitem{hashimoto2013field}
Kunimatsu Hashimoto, Fuminori Saito, Takashi Yamamoto, and Koichi Ikeda.
\newblock A field study of the human support robot in the home environment.
\newblock In {\em 2013 IEEE Workshop on Advanced Robotics and its Social
  Impacts}, pages 143--150. IEEE, 2013.

\bibitem{tanioka2019nursing}
Tetsuya Tanioka.
\newblock Nursing and rehabilitative care of the elderly using humanoid robots.
\newblock {\em The Journal of Medical Investigation}, 66(1.2):19--23, 2019.

\bibitem{tiagoPAL}
PAL Robotics.
\newblock Tiago.

\bibitem{juel2020smooth}
William~K Juel, Frederik Haarslev, Eduardo~R Ramirez, Emanuela Marchetti,
  Kerstin Fischer, Danish Shaikh, Poramate Manoonpong, Christian Hauch, Leon
  Bodenhagen, and Norbert Kr{\"u}ger.
\newblock Smooth robot: Design for a novel modular welfare robot.
\newblock {\em Journal of Intelligent \& Robotic Systems}, 98(1):19--37, 2020.

\bibitem{astroAMAZON}
Amazon.
\newblock Introducing amazon astro – household robot for home monitoring,
  with alexa, 2021.

\bibitem{toshev2014deeppose}
Alexander Toshev and Christian Szegedy.
\newblock Deeppose: Human pose estimation via deep neural networks.
\newblock In {\em Proceedings of the IEEE conference on computer vision and
  pattern recognition}, pages 1653--1660, 2014.

\bibitem{cao2019openpose}
Zhe Cao, Gines Hidalgo, Tomas Simon, Shih-En Wei, and Yaser Sheikh.
\newblock Openpose: realtime multi-person 2d pose estimation using part
  affinity fields.
\newblock {\em IEEE transactions on pattern analysis and machine intelligence},
  43(1):172--186, 2019.

\bibitem{zhao2019object}
Zhong-Qiu Zhao, Peng Zheng, Shou-tao Xu, and Xindong Wu.
\newblock Object detection with deep learning: A review.
\newblock {\em IEEE transactions on neural networks and learning systems},
  30(11):3212--3232, 2019.

\bibitem{deng2009imagenet}
Jia Deng, Wei Dong, Richard Socher, Li-Jia Li, Kai Li, and Li~Fei-Fei.
\newblock Imagenet: A large-scale hierarchical image database.
\newblock In {\em 2009 IEEE conference on computer vision and pattern
  recognition}, pages 248--255. Ieee, 2009.

\bibitem{alom2018history}
Md~Zahangir Alom, Tarek~M Taha, Christopher Yakopcic, Stefan Westberg, Paheding
  Sidike, Mst~Shamima Nasrin, Brian~C Van~Esesn, Abdul A~S Awwal, and Vijayan~K
  Asari.
\newblock The history began from alexnet: A comprehensive survey on deep
  learning approaches.
\newblock {\em arXiv preprint arXiv:1803.01164}, 2018.

\bibitem{mateus2019efficient}
Andre Mateus, David Ribeiro, Pedro Miraldo, and Jacinto~C Nascimento.
\newblock Efficient and robust pedestrian detection using deep learning for
  human-aware navigation.
\newblock {\em Robotics and Autonomous Systems}, 113:23--37, 2019.

\bibitem{redmon2016you}
Joseph Redmon, Santosh Divvala, Ross Girshick, and Ali Farhadi.
\newblock You only look once: Unified, real-time object detection.
\newblock In {\em Proceedings of the IEEE conference on computer vision and
  pattern recognition}, pages 779--788, 2016.

\bibitem{liu2016ssd}
Wei Liu, Dragomir Anguelov, Dumitru Erhan, Christian Szegedy, Scott Reed,
  Cheng-Yang Fu, and Alexander~C Berg.
\newblock Ssd: Single shot multibox detector.
\newblock In {\em European conference on computer vision}, pages 21--37.
  Springer, 2016.

\bibitem{PoseNet}
George Papandreou, Tyler Zhu, Liang{-}Chieh Chen, Spyros Gidaris, Jonathan
  Tompson, and Kevin Murphy.
\newblock Personlab: Person pose estimation and instance segmentation with a
  bottom-up, part-based, geometric embedding model.
\newblock {\em CoRR}, abs/1803.08225, 2018.

\bibitem{moore2015talking}
Roger~K. Moore.
\newblock {\em From Talking and Listening Robots to Intelligent Communicative
  Machines}, pages 317 -- 336.
\newblock WALTER DE GRUYTER Incorporated, Boston, MA, markowsky, judith
  edition, 2014.

\bibitem{skantze2021turn}
Gabriel Skantze.
\newblock Turn-taking in conversational systems and human-robot interaction: a
  review.
\newblock {\em Computer Speech \& Language}, 67:101178, 2021.

\bibitem{tenney2019bert}
Ian Tenney, Dipanjan Das, and Ellie Pavlick.
\newblock Bert rediscovers the classical nlp pipeline.
\newblock {\em arXiv preprint arXiv:1905.05950}, 2019.

\bibitem{TurtleBot3}
Turtlebot3 on robotis official site.
\newblock
  \url{https://emanual.robotis.com/docs/en/platform/turtlebot3/overview/}.

\bibitem{TurtleBot2}
Turtlebot2 on turtlebot official site.
\newblock \url{https://www.turtlebot.com/turtlebot2/}.

\bibitem{RosBot2Pro}
Rosbot2 pro on husarion official site.
\newblock \url{https://store.husarion.com/products/rosbot-pro}.

\bibitem{taheri2020OmniMobileRobot}
Hamid Taheri and Chun~Xia Zhao.
\newblock Omnidirectional mobile robots, mechanisms and navigation approaches.
\newblock {\em Mechanism and Machine Theory}, 153:103958, 2020.

\bibitem{ilon1975}
Bengt~Erland Ilon.
\newblock Wheels for a course stable selfpropelling vehicle movable in any
  desired direction on the ground or some other base, 1975.
\newblock US Patent 3,876,255.

\bibitem{pin1994}
Francois~G Pin and Stephen~M Killough.
\newblock A new family of omnidirectional and holonomic wheeled platforms for
  mobile robots.
\newblock {\em IEEE transactions on robotics and automation}, 10(4):480--489,
  1994.

\bibitem{salih2006}
Jefri Efendi~Mohd Salih, Mohamed Rizon, Sazali Yaacob, Abdul~Hamid Adom, and
  Mohd~Rozailan Mamat.
\newblock Designing omni-directional mobile robot with mecanum wheel [j.
\newblock In {\em American Journal of Applied Science s}. Citeseer, 2006.

\bibitem{cuevas2019}
Felizardo Cuevas, Oscar Castillo, and Prometeo Cortes-Antonio.
\newblock Towards an adaptive control strategy based on type-2 fuzzy logic for
  autonomous mobile robots.
\newblock In {\em 2019 IEEE International Conference on Fuzzy Systems
  (FUZZ-IEEE)}, pages 1--6. IEEE, 2019.

\bibitem{mourioux2006}
Gilles Mourioux, Cyril Novales, G{\'e}rard Poisson, and Pierre Vieyres.
\newblock Omni-directional robot with spherical orthogonal wheels: concepts and
  analyses.
\newblock In {\em Proceedings 2006 IEEE International Conference on Robotics
  and Automation, 2006. ICRA 2006.}, pages 3374--3379. IEEE, 2006.

\bibitem{tadakuma2007}
Kenjiro Tadakuma, Riichiro Tadakuma, and Jose Berengeres.
\newblock Development of holonomic omnidirectional vehicle with
  “omni-ball”: spherical wheels.
\newblock In {\em 2007 IEEE/RSJ International Conference on Intelligent Robots
  and Systems}, pages 33--39. IEEE, 2007.

\bibitem{ferriere1998}
Laurent Ferri{\`e}re and Beno{\^\i}t Raucent.
\newblock Rollmobs, a new universal wheel concept.
\newblock In {\em Proceedings. 1998 IEEE International Conference on Robotics
  and Automation (Cat. No. 98CH36146)}, volume~3, pages 1877--1882. IEEE, 1998.

\bibitem{ferland2010}
Fran{\c{c}}ois Ferland, Lionel Clavien, Julien Fr{\'e}my, Dominic
  L{\'e}tourneau, Fran{\c{c}}ois Michaud, and Michel Lauria.
\newblock Teleoperation of azimut-3, an omnidirectional non-holonomic platform
  with steerable wheels.
\newblock In {\em 2010 IEEE/RSJ International Conference on Intelligent Robots
  and Systems}, pages 2515--2516. IEEE, 2010.

\bibitem{Nexus4WD}
Nexus 4wd mecanum wheel mobile robot on nexus official site.
\newblock
  \url{https://www.nexusrobot.com/product/4wd-mecanum-wheel-mobile-arduino-robotics-car-10011.html}.

\bibitem{robotics7030047}
Olimpiya Saha and Prithviraj Dasgupta.
\newblock A comprehensive survey of recent trends in cloud robotics
  architectures and applications.
\newblock {\em Robotics}, 7(3), 2018.

\bibitem{ROS2}
Steven Macenski, Tully Foote, Brian Gerkey, Chris Lalancette, and William
  Woodall.
\newblock Robot operating system 2: Design, architecture, and uses in the wild.
\newblock {\em Science Robotics}, 7(66):eabm6074, 2022.

\bibitem{ROS}
The robot operating system official site.
\newblock \url{https://www.ros.org/}.

\bibitem{ExploringThePerformance}
Yuya Maruyama, Shinpei Kato, and Takuya Azumi.
\newblock Exploring the performance of ros2.
\newblock 2016.

\bibitem{ROS2vsROS1}
Changes between ros2 and ros1.
\newblock \url{https://design.ros2.org/articles/changes.html}.

\bibitem{zhang2018fast}
Xuetao Zhang, Zhenxue Chen, QM~Jonathan Wu, Lei Cai, Dan Lu, and Xianming Li.
\newblock Fast semantic segmentation for scene perception.
\newblock {\em IEEE Transactions on Industrial Informatics}, 15(2):1183--1192,
  2018.

\bibitem{Sort}
Alex Bewley, ZongYuan Ge, Lionel Ott, Fabio Ramos, and Ben Upcroft.
\newblock Simple online and realtime tracking.
\newblock {\em CoRR}, abs/1602.00763, 2016.

\bibitem{debeunne2020review}
C{\'e}sar Debeunne and Damien Vivet.
\newblock A review of visual-lidar fusion based simultaneous localization and
  mapping.
\newblock {\em Sensors}, 20(7):2068, 2020.

\bibitem{macenski2020marathon}
Steve Macenski, Francisco Mart{\'\i}n, Ruffin White, and Jonatan~Gin{\'e}s
  Clavero.
\newblock The marathon 2: A navigation system.
\newblock In {\em 2020 IEEE/RSJ International Conference on Intelligent Robots
  and Systems (IROS)}, pages 2718--2725. IEEE, 2020.

\bibitem{Slam_Toolbox}
Steve Macenski and Ivona Jambrecic.
\newblock Slam toolbox: Slam for the dynamic world.
\newblock {\em Journal of Open Source Software}, 6(61):2783, 2021.

\bibitem{berg2021keyword}
Axel Berg, Mark O'Connor, and Miguel~Tairum Cruz.
\newblock Keyword transformer: A self-attention model for keyword spotting.
\newblock {\em arXiv preprint arXiv:2104.00769}, 2021.

\bibitem{andreev2021speech}
Alexey Andreev and Kirill Chuvilin.
\newblock Speech recognition for mobile linux distrubitions in the case of
  aurora os.
\newblock In {\em 2021 29th Conference of Open Innovations Association
  (FRUCT)}, pages 14--21. IEEE, 2021.

\bibitem{dosovitskiy2020image}
Alexey Dosovitskiy, Lucas Beyer, Alexander Kolesnikov, Dirk Weissenborn,
  Xiaohua Zhai, Thomas Unterthiner, Mostafa Dehghani, Matthias Minderer, Georg
  Heigold, Sylvain Gelly, et~al.
\newblock An image is worth 16x16 words: Transformers for image recognition at
  scale.
\newblock {\em arXiv preprint arXiv:2010.11929}, 2020.

\bibitem{warden2018speech}
Pete Warden.
\newblock Speech commands: A dataset for limited-vocabulary speech recognition.
\newblock {\em arXiv preprint arXiv:1804.03209}, 2018.

\end{thebibliography}

\end{document}